\documentclass{article}

\usepackage[preprint]{neurips_2026}
\usepackage{microtype}
\usepackage{graphicx}
\usepackage{subcaption}
\usepackage{booktabs}
\usepackage{siunitx}
\usepackage{tikz}
\usetikzlibrary{positioning,calc,backgrounds,fit,shapes.geometric,patterns}
\usepackage{multirow}
\usepackage{wrapfig}
\usepackage{hyperref}

\usepackage{amsmath}
\usepackage{amssymb}
\usepackage{mathtools}
\usepackage{amsthm}
\usepackage{bm}
\usepackage{bbm}

\usepackage{algorithm}
\usepackage{algorithmic}

\usepackage{listings}
\usepackage{xcolor}

\usepackage[noabbrev]{cleveref}
\crefname{figure}{Fig.}{Figs.}
\Crefname{figure}{Figure}{Figures}
\crefname{equation}{Eq.}{Eqs.}
\Crefname{equation}{Equation}{Equations}
\crefname{table}{Table}{Tables}
\Crefname{table}{Table}{Tables}
\crefname{section}{Section}{Sections}
\Crefname{section}{Section}{Sections}
\crefname{algorithm}{Algorithm}{Algorithms}
\Crefname{algorithm}{Algorithm}{Algorithms}

\theoremstyle{plain}

\theoremstyle{definition}

\theoremstyle{remark}

\definecolor{xlinkcolor}{rgb}{0.7752941176470588, 0.22078431372549023, 0.2262745098039215}
\hypersetup{
  colorlinks=true,
  linkcolor=xlinkcolor,
  citecolor=xlinkcolor,
  urlcolor=xlinkcolor
}

\definecolor{codeblue}{RGB}{41,98,150}
\definecolor{codepurple}{RGB}{112,74,139}
\definecolor{codegreen}{RGB}{62,120,80}
\definecolor{codegray}{RGB}{110,110,110}
\definecolor{backcolour}{RGB}{248,248,248}
\definecolor{accentblue}{RGB}{70,130,180}

\lstdefinestyle{mystyle}{
  backgroundcolor=\color{backcolour},
  commentstyle=\itshape\color{codegray},
  keywordstyle=\color{codeblue},
  numberstyle=\tiny\color{codegray},
  stringstyle=\color{codegreen},
  basicstyle=\ttfamily\small,
  breakatwhitespace=false,
  breaklines=true,
  captionpos=b,
  keepspaces=true,
  showspaces=false,
  showstringspaces=false,
  showtabs=false,
  tabsize=4,
  frame=l,
  framesep=8pt,
  framerule=2pt,
  rulecolor=\color{accentblue},
  xleftmargin=12pt,
  aboveskip=10pt,
  belowskip=10pt,
}
\newcommand{\arcsecond}{\ensuremath{\,\mathrm{arcsec}}}

\DeclareMathOperator{\E}{\mathbb{E}}
\newcommand{\R}{\mathbb{R}}

\newcommand{\given}{\,|\,}

\newcommand{\params}{\theta}
\newcommand{\obs}{x}
\newcommand{\dd}{\mathrm{d}}

\title{Program Synthesis for Simulation-Based Inference: \\
Joint Model Selection and Parameter Estimation}

\author{%
  Siddharth Mishra-Sharma\thanks{Also at Anthropic. Work done at Boston University.} \\
  {Faculty of Computing \& Data Sciences} \\
  {Boston University} \\
  \texttt{smishras@bu.edu} \\
}

\begin{document}

\maketitle

\begin{abstract}
Neural simulation-based inference enables parameter estimation for complex models, but typically requires the user to specify a simulator encoding a fixed model structure.
We present a framework for joint model selection and parameter estimation that combines large language models for program synthesis with neural simulation-based inference.
Given a natural language description of the system and data under investigation, an LLM proposes candidate simulator programs which are iteratively refined via feedback-driven mutation and evaluated using neural density estimation.
The approach enables simulation-based inference over a pool of models, not just parameters within a fixed model.
On benchmarks spanning deterministic dynamics, stochastic epidemic models, and dark matter substructure inference from gravitational-lensing images, the method identifies plausible model families from open-ended prompts, with accuracy that reflects the information content of the data and identifiability of candidate models.
\end{abstract}

\section{Introduction}
\label{sec:intro}

A central challenge in scientific modeling is navigating the vast space of possible descriptions of an underlying system.
Traditionally, researchers manually propose candidate models based on domain expertise, sometimes comparing these models using statistical criteria such as Bayes factors.
This manual process is labor-intensive, potentially overlooking alternative, well-motivated model structures that better explain the data.
The space of scientific models is the space of possible compatible simulator programs, which is challenging to explore systematically.

The capabilities of large language models (LLMs) have been improving rapidly \citep{google2025gemini,anthropic2025claude,singh2025openai,guo2025deepseek,team2025kimi}, and LLMs today can reliably produce executable programs that implement differential equations, stochastic simulations, and other mechanistic models.
This suggests using LLMs as an \emph{implicit prior} over the space of scientific simulators: draw candidate programs from this prior, then filter them through Bayesian model comparison.

Separately, simulation-based inference (SBI) methods have emerged as powerful tools for Bayesian inference when likelihoods are intractable but forward simulations are available \citep{cranmer2020frontier,deistler2025simulation}.
As broad classes, neural posterior estimation (NPE) methods learn the posterior directly \citep{papamakarios2016fast,lueckmann2017flexible,greenberg2019automatic}, neural likelihood estimation (NLE) methods learn the likelihood function \citep{papamakarios2019sequential}, and neural ratio estimation (NRE) methods learn likelihood ratios or likelihood-to-evidence ratios \citep{cranmer2015approximating,hermans2020likelihood,miller2021truncated}.
These methods have found widespread application in physics, cosmology, neuroscience, ecology, and other fields where mechanistic simulators are available but likelihood functions are intractable \citep[see][for a curated collection of applications]{Mishra-Sharma_Awesome_Neural_SBI_2023}.

In this work, we unify these advances into a framework for joint model selection and parameter estimation.
Given a natural language description of a scientific problem and a dataset under consideration, the method proceeds by first using an LLM to synthesize multiple candidate simulator programs as executable code, each with specified parameter priors that reflect domain-appropriate ranges.
It then generates simulations from all candidate models under their respective priors, trains neural classifiers to estimate log Bayes factors for model comparison, and fits neural posterior estimators for each model to enable parameter inference.
Iterating over several rounds, the final output is a joint posterior $p(M, \params \given \obs)$ over both the discrete model space within the generated candidate pool and continuous parameter space associated with each model.

The framework handles a growing candidate pool via pairwise ratio estimation without retraining a joint model over all candidates, implements an Occam-style complexity penalty relative to the LLM-specified priors, and produces per-model parameter posteriors.
We validate on controlled benchmarks with known ground truth, spanning deterministic function-fitting, stochastic compartmental dynamics, and gravitational-lensing images.

\section{Related Work}
\label{sec:related}

\paragraph{Simulation-Based Inference}
Simulation-based inference methods enable inference in settings where the likelihood function is intractable or unavailable, but forward simulation from a mechanistic model is possible.
These approaches, also known as likelihood-free inference, have a long history in statistics under the umbrella of approximate Bayesian computation (ABC), which accepts parameter proposals based on the similarity between simulated and observed data \citep{rubin1984,sisson2018handbook}, with ABC-SMC methods (ABC combined with sequential Monte Carlo, SMC) extending to model comparison \citep{toni2009approximate}.
Neural approaches have substantially improved the efficiency and accuracy of SBI by training density estimators or classifiers to directly approximate posteriors, likelihoods, or likelihood ratios without necessarily relying on summary descriptions of high-dimensional data \citep{papamakarios2016fast,lueckmann2017flexible,greenberg2019automatic,papamakarios2019sequential,hermans2020likelihood,miller2021truncated,deistler2022truncated,cranmer2020frontier}.
Our work extends SBI to settings where the model space itself is generated rather than pre-specified.

\paragraph{Program Synthesis and LLM-Based Discovery}
Program synthesis aims to automatically generate programs satisfying a given specification \citep{gulwani2017program}.
LLMs have achieved strong results on synthesis benchmarks \citep{chen2021evaluating,austin2021program}.
Most relevant to our work are evolutionary approaches that use LLMs for scientific and algorithmic discovery.
FunSearch \citep{romera2024funsearch} pairs an LLM with an evaluator in an evolutionary loop and AlphaEvolve \citep{novikov2025alphaevolve} extends this to evolve entire codebases.
LLM-SR \citep{shojaee2024llmsr} combines LLMs with evolutionary search for symbolic equation discovery.
Our work shares the evolutionary LLM framework but differs in two key ways: we target scientific simulators rather than algorithms, and we perform Bayesian inference over the generated programs rather than optimizing a single objective.

\paragraph{LLM-Based Probabilistic Model Discovery}
The most closely related concurrent work is {ModelSMC} \citep{wahl2026modelsmc}, which independently casts LLM-based mechanistic model discovery as SMC with executable programs as particles, iteratively proposed and refined by an LLM.
Both methods marginalize parameters and produce model posteriors at the SMC step, with per-model parameter posteriors obtained downstream via neural posterior estimation. The substantive differences are in how the marginal likelihood is estimated and how pairwise comparisons are aggregated.
First, ModelSMC weights particles using neural likelihood estimation, specifically a pretrained tabular density estimator (TabPFN) applied to the per-model marginal $p(\obs \given m) = \int p(\params)\, p_\phi(\obs \given m, \params)\, \dd\params$, whereas we use neural ratio estimation \citep{jeffrey2024evidence}: pairwise classifiers trained with the exponential loss target log Bayes factors directly without estimating either likelihood.
Second, we aggregate the resulting pairwise log Bayes factors by projecting them onto a globally consistent log-evidence vector via least squares (\cref{eq:ls_aggregation}), which removes order-dependence and the implicit inverse-temperature distortion of naive averaging. ModelSMC additionally validates on real-world pharmacology and neuroscience recordings, whereas all our experiments use synthetic ground truth.
Other recent probabilistic-LLM approaches include Large Language Bayes \citep{domke2025large}, which generates probabilistic programs and performs likelihood-weighted model averaging on tractable likelihoods.

\section{Method}
\label{sec:method}

We present the framework for joint inference over model space $\mathcal{M}$ and parameter space $\Theta$.
The goal is to compute the joint posterior distribution over both discrete model indices and continuous model parameters associated with each model given observed data $\obs$:
\begin{equation}
    p(M, \params \given \obs) = p(\params \given \obs, M) \cdot p(M \given \obs),
\label{eq:joint}
\end{equation}
where $M \in \mathcal{M} = \{M_1, \ldots, M_K\}$ is a discrete model index, $\params \in \Theta_M$ are the continuous parameters specific to model $M$, and $\obs$ is the observed data.
The factorization in \cref{eq:joint} separates the problem into model selection, computing $p(M \given \obs)$, and parameter inference conditional on each model, computing $p(\params \given \obs, M)$.
We address both components using neural density estimation.

\subsection{LLM-Based Program Synthesis}

The first stage uses a large language model to generate plausible candidate simulator programs from a natural language problem description.
Given a problem description $\mathcal{P}$ specifying the phenomenon to be modeled, the observable quantities, and any relevant domain constraints, we prompt the LLM to generate $K$ diverse candidate models:
\begin{equation}
    \{(M_k, \pi_k)\}_{k=1}^K \sim p_{\text{LLM}}(\cdot \given \mathcal{P}),
\end{equation}
where each $M_k$ is executable code defining a simulator function and $\pi_k$ specifies the prior distribution over the parameters of that model.

Each simulator accepts a parameter dictionary, simulates forward (numerical integration, stochastic sampling, etc.), and returns the observables as a numerical array.
The LLM is prompted to generate diverse candidates exploring different modeling assumptions; we provide the full prompt template in Appendix~\ref{app:prompts}.

Before inference we validate each program by parsing its code and executing it on random parameter draws, checking that the output is finite; failing programs are discarded and the round proceeds with the surviving candidates (simulations with unexpected shape are additionally dropped when training data is generated). For the subhalo benchmark the LLM emits parameter configurations for a fixed simulator rather than code, so this validation step does not apply.

\subsection{Sequential Model Refinement}

Rather than treating program synthesis as a one-shot generation, we employ an iterative refinement strategy inspired by SMC that progressively focuses on promising regions of model space.
This refinement approach draws inspiration from sequential NPE methods \citep{greenberg2019automatic}, but operates over the discrete space of programs rather than continuous parameters.

In round 1, we generate $K$ diverse candidates from the LLM prior $p_{\text{LLM}}(\cdot \given \mathcal{P})$ and compute importance weights $w_k \propto p(\obs \given M_k)$ via ratio estimation (Section~\ref{sec:ratio}).
In subsequent rounds, we resample models proportional to their weights (with replacement), then use the LLM as a {mutation kernel} to generate variations of the selected models:
\begin{equation}
    M'_k \sim p_{\text{LLM}}(\cdot \given M_k, \mathcal{P}, \text{feedback}),
\end{equation}
where the mutation prompt includes the code and description of the parent model, encouraging the LLM to propose structurally related variants.
The \emph{feedback} term summarises the parent's standing in the current pool: its posterior weight together with the names and weights of the top- and bottom-ranked models in the round.
This grounds the proposal in the competitive landscape rather than in the parent alone; in the subhalo benchmark, parents with low weight are additionally nudged toward more substantial structural changes, while high-weight parents are pointed toward nearby variations of the currently dominant region of model space.
Benchmark-specific feedback augmentations are noted in Section~\ref{sec:experiments} when discussing the respective experiments.

Resampling concentrates computational resources on promising regions of program space, while the mutation kernel lets the LLM propose structurally related variants unlikely under independent sampling.
The effective sample size (ESS) of the weight vector provides a natural diagnostic: low ESS means weight has concentrated on a single model, while high ESS means competitive alternatives remain.
The loop terminates at the earlier of a fixed budget of $R$ rounds or when a single model reaches posterior weight $> 0.9$; in practice, clear-signal benchmarks converge in 1--5 rounds and ambiguous ones run to the budget while retaining competitive alternative models.

\paragraph{Theoretical Considerations}
This procedure is best understood as population-based evolutionary search guided by evidence estimates, not a rigorous SMC sampler targeting $p(M \given \obs)$.
Standard SMC requires either that the mutation kernel satisfy detailed balance, or that importance weights correct for the proposal density $q(M' \given M)$; the LLM mutation kernel is neither symmetric nor tractable to evaluate, precluding either correction.
Instead, resampling concentrates computational resources on promising models while LLM mutation explores structurally related variants analogous to adaptive importance sampling, where the proposal is iteratively refined but the final weights are computed from a fresh evaluation rather than from the proposal trajectory.
The refinement rounds thus serve to improve the candidate pool; the \emph{final} evaluation pool is subject to Bayesian model comparison.
Accordingly, the final model weights, derived from ratio estimation at the last round, are Bayesian posterior probabilities \emph{conditional on the final evaluation pool}: they reflect the relative marginal likelihoods (up to ratio-estimation error) of the models actually compared in the last round, rather than a posterior over the full space of programs.

\subsection{Neural Ratio Estimation for Model Comparison}
\label{sec:ratio}

The Bayes factor between two models $M_i$ and $M_j$ is the ratio of their marginal likelihoods:
\begin{equation}
    \text{BF}_{ij} = \frac{p(\obs \given M_i)}{p(\obs \given M_j)} = \frac{\int p(\obs \given \params, M_i) \, p(\params \given M_i) \, \dd\params}{\int p(\obs \given \params, M_j) \, p(\params \given M_j) \, \dd\params}.
\end{equation}
Computing these integrals directly is intractable, but they encode the desirable property of \emph{Bayesian Occam's razor}, i.e. models with more parameters or wider priors must ``spread'' their prior probability mass over a larger volume of parameter space, reducing their marginal likelihood even if they can achieve the same peak likelihood as a simpler model.
This automatic complexity penalty is essential for principled model comparison.

\paragraph{Evidence Networks}
Standard cross-entropy classification learns a density ratio that, with balanced training and equal model priors, can in principle yield the Bayes factor \citep{cranmer2015approximating,hermans2020likelihood}.
However, this requires careful calibration, and cross-entropy classifiers tend to saturate for extreme likelihood ratios.
We instead employ \emph{evidence networks} \citep{jeffrey2024evidence}, whose exponential loss is designed to target $\log \text{BF}$ more directly and empirically provides better-behaved estimates across a wide range of evidence strengths.

For each pair of models $(M_i, M_j)$, we generate simulations from both models under their respective priors and train a neural network $f_\phi: \obs \mapsto \R$ using the exponential loss:
\begin{equation}
    \mathcal{L}(f) = \E_{m, \obs} \left[ \exp\left( (0.5 - m) \cdot f(\obs) \right) \right],
\label{eq:exp_loss}
\end{equation}
where $m \in \{0, 1\}$ indicates which model generated the simulation ($m=0$ for $M_i$, $m=1$ for $M_j$), and the expectation is over balanced samples from both models.

Taking the functional derivative of \cref{eq:exp_loss} with respect to $f$ and setting it to zero gives
\begin{equation}
    p(\obs|M_i) e^{0.5 f^*} = p(\obs|M_j) e^{-0.5 f^*}.
\end{equation}
Solving for $f^*$ yields $f^*(\obs) = \log(p(\obs|M_j)/p(\obs|M_i)) \equiv \log \text{BF}_{ji}$.

Training samples are generated by first drawing $\params \sim \pi_k$ and then simulating $\obs \sim p(\cdot|\params, M_k)$, so they follow the prior-marginalized distribution $p(\obs|M_k) = \int p(\obs|\params, M_k) \pi_k(\params) \, \dd\params$; the optimum $f^*$ is therefore the log ratio of \emph{marginal} likelihoods, the log Bayes factor, read directly from the network output.

When log Bayes factors are large, $\exp((0.5{-}m) f)$ overflows numerically.
Following \citet{jeffrey2024evidence}, we train with the $\ell$-POP transform $J_2(f) = f(1{+}|f|)$, replacing $f$ in \cref{eq:exp_loss} by $J_2(f)$.
The resulting optimum satisfies $J_2(f^*) = \log \mathrm{BF}_{ji}$, so the log Bayes factor is recovered by applying $J_2$ to the trained network's output.

\paragraph{Model Posterior from Pairwise Comparisons}
We train $\binom{K}{2}$ pairwise ratio estimators, yielding estimates $\hat{s}_{ij} \approx \log \text{BF}_{ij}$.
A natural way to turn these into a model posterior is to average the pairwise comparisons for each model, $\log p(M_k \given \obs) \propto \frac{1}{K-1} \sum_{j \neq k} \hat{s}_{kj}$. This is subtly incorrect, because the average compares each model against a baseline that excludes the model itself. With exact estimates $\hat{s}_{kj} = \ell_k - \ell_j$, the average equals $\ell_k - \bar\ell_{\setminus k}$, where $\bar\ell_{\setminus k}$ is the mean log-evidence of the \emph{other} $K{-}1$ models; rewriting gives $\tfrac{K}{K-1}(\ell_k - \bar\ell)$, so the self-excluding baseline inflates every centered log-evidence by $K/(K{-}1)$. After the softmax that defines the posterior, this factor acts as an inverse temperature $\beta > 1$ and the resulting distribution over models is \emph{sharper} than the true Bayesian posterior. Averaging therefore systematically overstates the evidence for the leading model; we derive this in Appendix~\ref{app:ls_aggregation}.
A second problem is non-transitivity. Because each pairwise estimator is trained independently on a finite simulation budget, the collection of estimates need not be consistent with any single log-evidence vector: it is possible to observe $\hat{s}_{AB} > 0$ and $\hat{s}_{BC} > 0$ while $\hat{s}_{AC} < 0$. Averaging doesn't account for this, so cyclic disagreement among the estimators is silently folded into the posterior.

We instead seek log-evidences $\ell_k \approx \log p(\obs \given M_k)$ that best explain all $\binom{K}{2}$ pairwise observations simultaneously in a least-squares sense:
\begin{equation}
    \min_{\boldsymbol{\ell}} \sum_{i < j} \left( \hat{s}_{ij} - (\ell_i - \ell_j) \right)^2, \qquad \text{subject to } \textstyle\sum_k \ell_k = 0.
    \label{eq:ls_aggregation}
\end{equation}
This is the maximum-likelihood estimate of the log-evidences under i.i.d.\ Gaussian noise on the pairwise estimates; Appendix~\ref{app:ls_aggregation} gives the closed-form solution and shows that the fit residual serves as a diagnostic for cyclic disagreement among the estimators.
The model posterior is then $p(M_k \given \obs) \propto \exp(\ell_k) \, p(M_k)$.

\subsection{Neural Posterior Estimation for Parameters}

For each candidate model $M_k$, we train a neural posterior estimator to approximate the conditional parameter posterior $p(\params \given \obs, M_k)$.
We use masked autoregressive flows (MAFs) as the density estimator \citep{papamakarios2017masked}.

Training proceeds by generating simulation pairs $\{(\params^{(n)}, \obs^{(n)})\}_{n=1}^{N_{\rm sim}}$ where parameters are sampled from the prior $\params^{(n)} \sim \pi_k$ and observations are generated by running the simulator $\obs^{(n)} \sim p(\cdot \given \params^{(n)}, M_k)$.
The flow parameters $\psi$ are optimized to maximize the log-likelihood of the training parameters under the flow: $\mathcal{L}(\psi) = \frac{1}{N_{\rm sim}} \sum_{n=1}^{N_{\rm sim}} \log q_\psi(\params^{(n)} \given \obs^{(n)})$.
At inference time, given the actual observed data $\obs_{\text{obs}}$, we can sample from the approximate posterior $q_\psi(\params \given \obs_{\text{obs}})$ or evaluate its density at any parameter value.

\subsection{Joint Posterior and Inference}

Given $p(M \given \obs)$ from ratio estimation and NPE posteriors $q_{\psi_k}(\params \given \obs)$ for the top-ranked surviving models, samples from the joint posterior are obtained by ancestral sampling: draw $M \sim p(M \given \obs)$, then $\params \sim q_{\psi_M}(\params \given \obs)$.
This supports standard downstream operations like model-averaged predictions and maximum a posteriori (MAP) model identification along with its parameter posterior.
Algorithm~\ref{alg:main} summarizes the procedure.

\begin{algorithm}[H]
\caption{Evolutionary Search over Program Space}
\label{alg:main}
\begin{algorithmic}[1]
\REQUIRE Problem description $\mathcal{P}$, observation $\obs$, particles $K$, rounds $R$
\STATE \textbf{// Round 1: Generate from LLM prior}
\STATE $\{M_k\}_{k=1}^K \gets \text{LLM}(\mathcal{P})$; validate programs
\STATE Generate simulations, train classifiers, compute weights $w_k$
\FOR{$r = 2$ to $R$}
    \STATE \textbf{// Resample proportional to weights}
    \STATE $\{\text{parent}_k\}_{k=1}^K \gets \text{Multinomial}(\{w_k\})$
    \STATE \textbf{// Mutate via LLM}
    \FOR{$k = 1$ to $K$}
        \STATE $M_k \gets \text{LLM}(\text{parent}_k, \mathcal{P}, \text{``generate variation''})$
    \ENDFOR
    \STATE Validate, simulate, compute new weights $w_k$
    \STATE Compute ESS $= 1/\sum_k \bar{w}_k^2$ (normalized weights); check convergence
\ENDFOR
\STATE \textbf{// Final inference}
\STATE Normalize weights to get $p(M_k \given \obs)$ within explored pool
\STATE Train NPE flows $q_{\psi_k}(\params \given \obs)$ for the top-ranked surviving models by final weight (with $w_k > 0.01$)
\STATE \textbf{return} $p(M \given \obs)$, $\{q_{\psi_k}\}$
\end{algorithmic}
\end{algorithm}

\section{Experiments}
\label{sec:experiments}

We evaluate on three controlled benchmarks with known ground truth: a damped harmonic oscillator (essentially function fitting), SIR compartmental dynamics with three ground-truth variants (stochastic latent structure; related-family discrimination), and dark matter subhalo population models inferred from simulated gravitational-lensing images.

\subsection{Experimental Setup}

For program synthesis, we use Claude Opus 4.6 as the proposal model, with temperature $T=0.7$.
The prompt specifies the problem domain and observable quantities and requests diverse modeling approaches, but does {not} name specific model families or specify which models to generate. It also asks for a range of candidate complexities, including at least one minimal model, with a stated preference for simplicity; this biases which programs are \emph{proposed}, not how they are scored (full template in Appendix~\ref{app:prompts}).
The LLM also specifies parameter priors based on the problem description.
Coupling the prior to the simulator is natural in our setting: the simulator definition and the parameterization it induces are inseparable, so a candidate model is really a (program, prior) pair drawn jointly from $p_{\text{LLM}}$.
Poor priors are penalised by the marginal-likelihood machinery: priors that are too narrow miss parameter regions that fit the data, while priors that are too wide spread mass over implausible regions.
A model whose prior is badly mis-specified therefore loses weight to competitors with better-matched priors, and the refinement loop has the opportunity to propose revised priors alongside revised structure.

The prompt requires simulators to include observational noise matching the data-generating process, as the ratio estimator compares the full generative model including the noise process.
We use $K=8$ candidates per round with up to $R=10$ refinement rounds for ordinary differential equation (ODE) benchmarks (four of the ten SIR seeds ran with a five-round budget) and $R=15$ for the subhalo benchmark.
All experiments use 10 random seeds for ODE benchmarks and 3 seeds for the subhalo benchmark.
For the ODE benchmarks, whose observations are 1D time series, the mutation feedback (Section~\ref{sec:method}) additionally includes each candidate's root-mean-square error (RMSE) against the observation and the sign of its residual trend over time, together with a directive to revise the parent in the indicated direction (e.g.\ a candidate whose residuals trend upward is told to ``consider adding terms that increase predictions at larger values'').

For ratio estimation towards evidence computation, we train ensembles of 5 evidence networks per model pair, each a 4-layer multi-layer perceptron (MLP) with 256 hidden units and Gaussian error linear unit (GELU) activations, with 80/20 train/validation split and early stopping.
Training uses the exponential loss with $\ell$-POP transform described in Section~\ref{sec:ratio} and the Adam optimizer \citep{kingma2015adam} with learning rate $10^{-3}$.
Pairwise log Bayes factors are aggregated via least-squares projection onto globally consistent log-evidences (\cref{eq:ls_aggregation}).

For neural posterior estimation, we use MAFs implemented in the \texttt{sbi} package \citep{tejero2020sbi} to estimate $p(\params \given \obs, M)$ for the top-ranked surviving models.

The ODE benchmarks use 2{,}000 simulations per model per round; the subhalo benchmark uses 320 four-image datasets per model and four-image set-level evaluation.
ODE-benchmark runs take $\sim$2--30 minutes on a single GPU; subhalo-benchmark runs take $\sim$45--70 minutes per ground-truth/seed pair, dominated by per-round retraining of the set embedder.

\subsection{Benchmark 1: Damped Harmonic Oscillator}

As a warm-up, we generate observations from a damped cosine $x(t) = A e^{-\gamma t} \cos(\omega t) + \epsilon$, $\epsilon \sim \mathcal{N}(0, \sigma^2)$, with $A = 2.0$, $\gamma = 0.3$, $\omega = 3.0$, sampled at 50 uniform time points over $t \in [0, 10]$ with $\sigma = 0.1$. This is essentially a function-fitting / symbolic regression exercise.
The LLM prompt describes qualitative behavior (oscillation with decay) and the observation time span, without naming specific functional forms.
Typical candidates include damped sinusoids, Gaussian envelopes, power-law/algebraic decay, chirped oscillators, and two-mode beating models.

\begin{figure}[t]
\centering
\includegraphics[width=\textwidth]{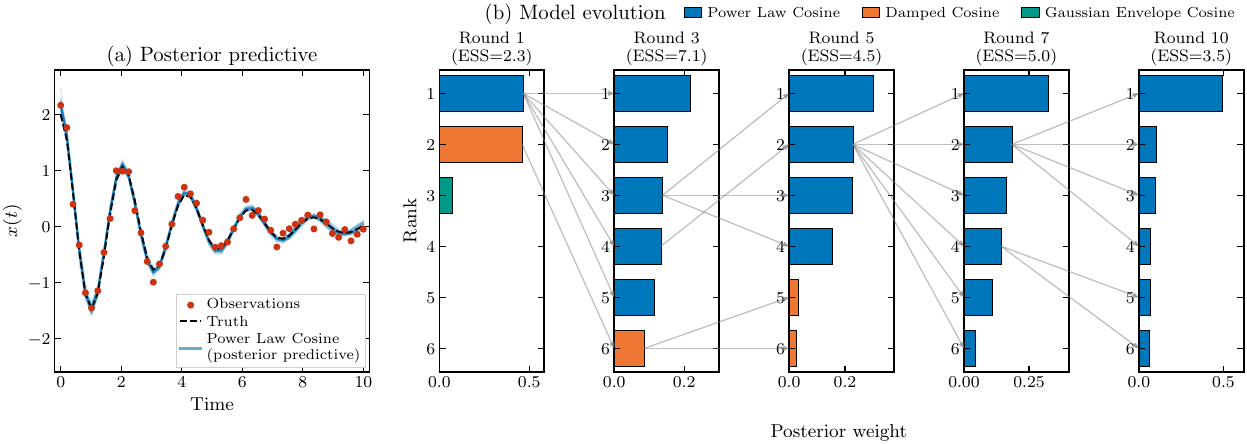}
\caption{Damped oscillator benchmark: the method recovers a damped, oscillating model and concentrates the posterior on it over successive SMC rounds. \emph{(a)} Posterior-predictive check of the winning LLM-generated program (a power-law-decay cosine variant): parameter-posterior draws (\textcolor[HTML]{0077BB}{blue} cloud) tightly track the \textcolor[HTML]{CC3311}{red} observations and the dashed ground truth. The posterior is computed by exact-likelihood MCMC on the stored program under the known Gaussian noise model, conditioned on the run's observation. \emph{(b)} Evolution of the model population across rounds, showing the top-6 models by posterior weight each round (bar color = LLM-proposed base family; ESS in the column title; gray arrows trace mutation lineage). Two families share the weight in round~1 and compete through the middle rounds; by round~10 the posterior has collapsed onto mutation variants of the winning power-law-decay cosine family.}
\label{fig:damped}
\end{figure}

Across 10 random seeds, the method identifies a model with both damping and oscillation in 10/10 cases.
The winning family varies across seeds (damped cosine, Gaussian-envelope cosine, and power-law-decay cosine each win in a comparable number of seeds), reflecting the diversity of the initial LLM-generated candidate pool (example generated programs in Appendix~\ref{app:models}); all correctly capture oscillation with decay.
Five seeds converge in round 1 and five run all 10 rounds, with ESS fluctuating between ${\sim}1$ and $8$ as within-family variants are injected and winnowed.
\Cref{fig:damped}\emph{(a)} shows posterior-predictive samples under the winning program closely tracking the ground truth, while \cref{fig:damped}\emph{(b)} visualizes the model-pool evolution across SMC rounds; the NPE machinery used for per-model parameter posteriors is approximately calibrated in the well-specified case of this benchmark's base model, within calibration uncertainty (see Appendix~\ref{app:sbc}).

\subsection{Benchmark 2: SIR Epidemic Dynamics}

This benchmark tests the method on a stochastic simulator with latent dynamics, and on multiple ground-truth variants within the same family.
We generate observations via Gillespie simulation of a susceptible--infected--recovered (SIR) population of size $N_{\rm pop} = 1000$: infection events $S \to I$ occur at rate $\beta S I / N_{\rm pop}$ and recovery events $I \to R$ at rate $\gamma I$, with true parameters $\beta = 0.4$, $\gamma = 0.1$ and initial state $(S_0, I_0, R_0) = (N_{\rm pop}{-}10,\,10,\,0)$. The observation is the infected count $I(t)$ evaluated at $10$ uniformly spaced times over $t \in [0, 100]$ with additive Gaussian noise of standard deviation $\sigma_{\text{obs}} = 5$ (clipped to be non-negative for the SIR ground truth).
Ground truths are SIR (standard), SEIR (added exposed compartment), and SIS (no immunity, endemic equilibrium); the prompt asks for diverse mechanistic epidemic simulators and lets the LLM choose the structure (compartments, differential equations, stochastic jump processes, or hybrids), without naming specific formulations.

\begin{figure}[t]
\centering
\includegraphics[width=\textwidth]{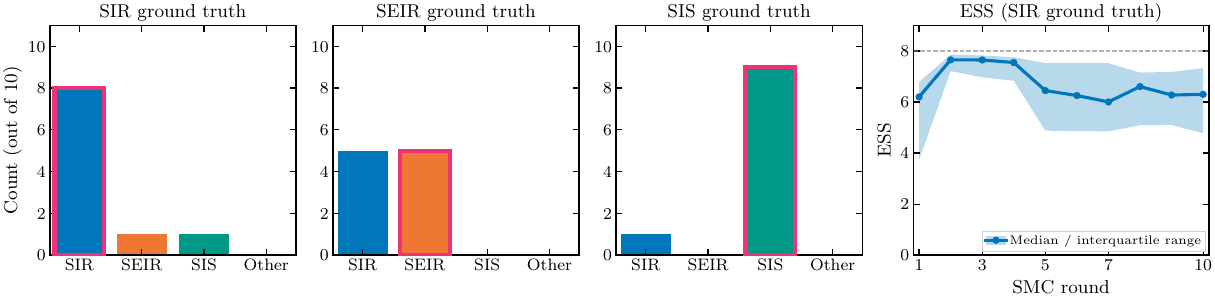}
\caption{SIR epidemic model selection: the method recovers the correct compartmental family when the dynamics are distinctive, and otherwise spreads weight across indistinguishable variants. \emph{Panels 1--3:} one per ground truth (SIR, SEIR, SIS); bar height is the number of seeds (of 10) whose MAP model falls in each family (\textcolor[HTML]{0077BB}{SIR}, \textcolor[HTML]{EE7733}{SEIR}, \textcolor[HTML]{009988}{SIS}, or Other), with the true family outlined in \textcolor[HTML]{EE3377}{magenta}. SIR is recovered in 8/10 seeds and SIS in 9/10; SEIR succeeds in only 5/10, because at this noise level it is observationally close to SIR and the Occam-style penalty favors the simpler family. \emph{Rightmost panel:} ESS of the $K{=}8$ pool across SMC rounds for the SIR ground truth (median and interquartile band over 10 seeds). The median ESS climbs to ${\sim}7.7$ by round~2 and stays in the ${\sim}6$--$7.5$ range thereafter: the posterior keeps mass on the various compartmental variants the data cannot separate.}
\label{fig:stochastic}
\end{figure}

\Cref{fig:stochastic} shows model selection results across the three SIR ground truths (10 seeds each).
Per expectation, model selection accuracy is highest for qualitatively distinct dynamics: SIS achieves 9/10 because its endemic equilibrium (no decline phase) is unmistakable, and SIR achieves 8/10.
SEIR is hardest at 5/10: in the 5 successful seeds SEIR wins with the posterior placing all of its mass on the SEIR family, while in the other 5 seeds SIR variants dominate instead. SEIR candidates are present in every initial pool, so the failure is not a proposal-diversity issue: SEIR and SIR produce qualitatively similar epidemic curves at these parameter values (the exposed compartment adds only a short time lag), and with 10 noisy observations the ratio estimator often cannot distinguish them, preferring the simpler SIR family via the implicit complexity penalty (Section~\ref{sec:ratio}).

The ESS of the model-weight vector is a natural diagnostic for whether the data distinguishes the candidates: it tends to $1$ when a single model dominates and to $K$ when weights spread uniformly.
The rightmost panel of \cref{fig:stochastic} shows ESS trajectories across 10 SIR seeds: the median climbs to ${\sim}7.7$ by round~2 and remains between ${\sim}6$ and $7.5$ thereafter, meaning the posterior consistently spreads across near-equivalent compartmental variants that the data cannot distinguish.

\subsection{Benchmark 3: Dark Matter Subhalo Populations}

Strong gravitational lensing of distant galaxies is an active observational probe of dark-matter substructure~\citep{vegetti2024strong}: when a foreground galaxy lenses a background source into arcs around its Einstein ring, each $\sim 10^{7\text{--}10}\,M_\odot$ subhalo in the lens adds a small perturbation to the deflection field, imprinting localized brightness anomalies on the arcs. Different dark-matter theories predict qualitatively different subhalo populations: cold dark matter (CDM) predicts abundant substructure with centrally concentrated (``cuspy'') Navarro--Frenk--White (NFW) density profiles~\citep{navarro1997universal}, while warm and self-interacting dark matter (WDM, SIDM) predict fewer subhalos with flattened central densities (``cored'' profiles). Inferring subhalo populations from samples of lensing images is an established target for simulation-based inference~\citep[e.g.,][]{brehmer2019mining,wagnercarena2024strong}. Given $N$ lensing systems hypothesized to share a common substructure population, we would like the inference machinery to identify the underlying family and, when the data permits, constrain its population-level parameters from a handful of noisy images, using a prompt that does not name any of the candidate theories.

We render 64$\times$64 pixel images (${\sim}0.1\arcsecond$ pixel scale) of a main lens populated with a Poisson realization of subhalos drawn from a power-law mass function over $[10^7, 10^{10}]\,M_\odot$; the abundance, mass-function slope, spatial scale, concentration, and density profile are population-level parameters ($\dd N/\dd M \propto M^{-1.9}$ for the CDM ground truth). Per image we vary the main-lens ellipticity ($q\in[0.7,0.95]$), Einstein radius ($\theta_E\in[1.3,1.8]\arcsecond$), and source offset; images are normalized to unit peak surface brightness, and per-pixel Gaussian noise $\sigma=0.005$ (peak signal-to-noise ${\sim}200$) is then added. Each evaluation presents a set of $N=4$ images sharing one population-level parameter vector $\theta^{*}$, treated hierarchically as described below. The LLM prompt describes only the physics and the simulator's parameter interface (subhalo count, mass-function slope, spatial scale, concentration, profile, core fraction) and emits $K=8$ candidate \emph{(description, prior)} pairs per round, anchored by one force-included smooth-null candidate. This benchmark is thus structured family selection over a fixed simulator interface, rather than open-ended synthesis of simulator code. The convolutional neural network (CNN) sees two channels per image. Channel~1 is the noisy observed image. Channel~2 is a noise-free residual map: the full image minus a smooth-lens-only rendering at the same (ground-truth) main-lens and source nuisance parameters, isolating the subhalo-induced perturbation. Supplying the ground-truth smooth model is an oracle simplification of a real analysis, in which the smooth lens would be inferred jointly and the subtraction would carry additional uncertainty into the residual channel.

\paragraph{Set-Level Evidence Network}
\label{sec:set-level}
Each lens in a set is drawn from a common population-level parameter $\theta^*$ but has per-system nuisance parameters (the main-lens and source geometry above), and the population is what we care about. Marginalizing over the shared $\theta$ gives
\begin{equation}
    p(\obs_{1:N} \given M) = \int \prod_{i=1}^N p(\obs_i \given \params, M)\, \pi(\params \given M)\, \dd\params,
\end{equation}
which is \emph{not} the product of per-system marginals; treating the $N$ observations as independent (summing per-image log BFs) would overstate evidence, while averaging would undercount it by a factor $N$. Both heuristics give wrong answers.

We handle this by training a single set-level evidence network on the joint distribution over $N$-tuples. Training samples are drawn hierarchically: $\theta \sim \pi(\theta \given M)$, then $\obs_i \sim p(\cdot\,|\,\theta, M)$ independently for $i=1,\dots,N$. A permutation-invariant DeepSets~\citep{zaheer2017deep} encoder maps the $N$-tuple to a single set embedding: a per-image feature network (a ResNet-18~\citep{he2016deep} on 2-channel image+residual input) produces per-image features, a shared MLP $\phi$ lifts them into the set-encoder representation, a mean-pool across the $N$ elements provides permutation invariance, and a post-pooling MLP $\rho$ (with $\log N$ concatenated as input, so the network can condition on dataset size) produces the final 128-dimensional set embedding. The pairwise evidence networks of Section~\ref{sec:ratio} then operate on these set embeddings unchanged, preserving the exponential-loss/$\ell$-POP machinery and targeting the hierarchical marginal likelihood. The set embedder is retrained at each SMC round so its feature space tracks the current candidate pool; the loop runs up to $R=15$ rounds.

\begin{figure}[t]
\centering
\includegraphics[width=0.98\columnwidth]{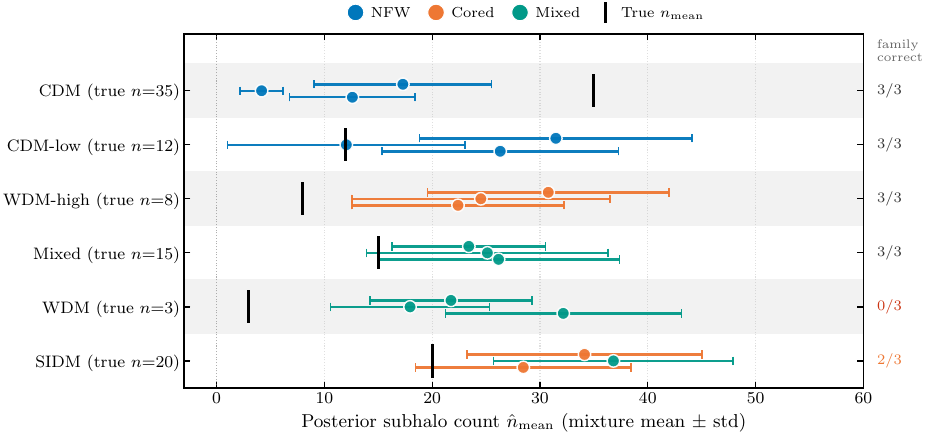}
\caption{Subhalo benchmark (4-lens set-level inference, 3 seeds per ground truth): the profile family is identified reliably (right margin: correct-family tally per ground truth, 14/18 overall), while the subhalo count is only weakly constrained at $N=4$. Each point is one seed's posterior mean subhalo count $\hat{n}_{\rm mean}$ (error bar: posterior standard deviation), colored by the family of its MAP model (\textcolor[HTML]{0077BB}{NFW}, \textcolor[HTML]{EE7733}{cored}, \textcolor[HTML]{009988}{mixed}); black ticks mark the true $n_{\rm mean}$. Count estimates regress toward the interior of the candidate priors: CDM undershoots its true $n_{\rm mean}=35$ and the sparse truths (WDM-high, WDM) overshoot, an information-content effect dissected in \cref{fig:subhalo_N_scaling}.}
\label{fig:subhalo_posterior}
\end{figure}

\paragraph{Substructure Family Identification}
The primary question is whether the framework can identify the underlying dark-matter profile family. We test six physically motivated ground truths spanning abundant-cuspy to sparse-cored populations, at three seeds each: CDM (NFW, $n_{\rm mean}{=}35$), CDM-low (NFW, $12$), WDM-high (cored, $8$), Mixed (50/50 NFW/cored, $15$), SIDM (cored, $20$), and WDM (cored, $3$). The framework recovers the correct family in 14/18 runs (\cref{fig:subhalo_posterior}, right margin): CDM, CDM-low, WDM-high, and Mixed at all seeds; SIDM at 2/3; WDM at 0/3, all three seeds converging on a \texttt{mixed}-profile candidate rather than pure \texttt{cored}, a near-degenerate but wrong call that we count as a failure. (With three seeds per ground truth the per-family rates carry wide uncertainties; the aggregate 14/18 has a 95\% Wilson interval of $[0.55, 0.91]$.) The subhalo count $n_{\rm mean}$ is a different story: at $N=4$ images it is only weakly identified. Count posteriors are broad (standard deviations up to ${\sim}13$ subhalos), and their means land closer to the interior of the candidates' prior ranges than to the truth: CDM \emph{undershoots} ($\hat n_{\rm mean}\in[4, 17]$ vs.\ true $35$), while the sparse truths \emph{overshoot} (WDM-high: $[22, 31]$ vs.\ true $8$; WDM: $[18, 32]$ vs.\ true $3$). The rest of this subsection traces both biases to a single cause, namely that four images at this resolution pin down the profile family but carry little information about the count, and then verifies that explanation by increasing $N$.

\paragraph{Origin of the Count Biases and Family Misidentification}
Both count biases follow from how much subhalo signal survives at the chosen image resolution. The flux perturbation from an NFW subhalo of mass $M$ has angular scale set by its Einstein radius, $\theta_{\rm sub} \approx 0.03\arcsecond\,(M / 10^7\,M_\odot)^{1/2}$. Subhalos below $10^8\,M_\odot$ therefore subtend less than the ${\sim}0.1\arcsecond$ pixel, and their imprint on the residual channel sits far below the image channel's per-pixel noise $\sigma$. With a $M^{-1.9}$ mass function, only ${\sim}12\%$ of CDM's 35 subhalos clear this threshold. Cored profiles behave differently: their flat centers spread each perturbation over many pixels. Direct measurement confirms the resulting asymmetry. Over 200 matched-geometry renderings per model, CDM ($n=35$, NFW) leaves a mean root-mean-square (RMS) residual of only ${\sim}0.5\,\sigma$ relative to the smooth reference, whereas WDM ($n=3$, cored) leaves ${\sim}55\,\sigma$.

This asymmetry predicts both directions of the count bias. On NFW truths, candidates with very different $n_{\rm mean}$ priors are likelihood-tied, so prior volume arbitrates and Occam selects the narrowest, sparsest candidate: CDM undershoots. (Minimal candidates are guaranteed present in every pool by the generation prompt; Appendix~\ref{app:prompts}.) On cored truths, the family is unambiguous, but four images say little about how the perturbation splits between subhalo count and per-subhalo structure (core radius, concentration, mass-function slope). The count posterior therefore stays broad and prior-driven, consistent with the prior-like widths in \cref{fig:subhalo_posterior}: the sparse truths overshoot.

\begin{figure}[t]
\centering
\includegraphics[width=\textwidth]{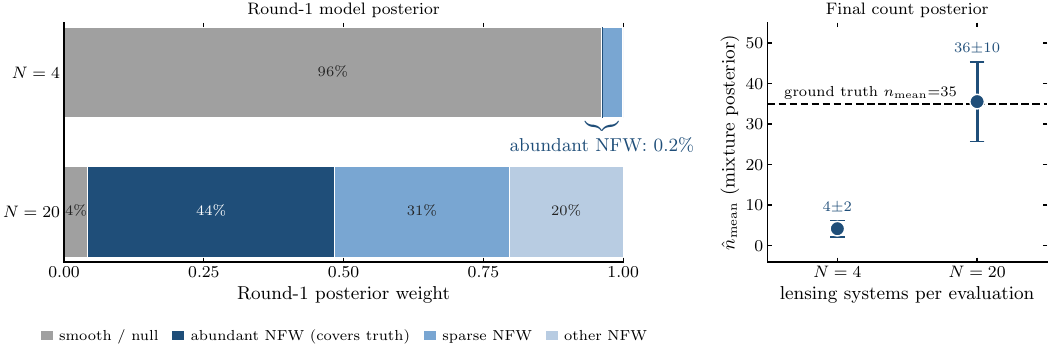}
\caption{The CDM count bias is a limit of data information content: adding lensing systems removes it. \emph{Left:} Round-1 posterior weight by candidate class as the number of lensing systems per evaluation $N$ grows. At $N=4$ the data cannot resolve the NFW substructure, so nearly all weight sits on the smooth-null candidate; at $N=20$ the weight shifts to the candidate covering the true abundant-NFW population. Wrong-family (cored/mixed) candidates receive negligible weight throughout and are omitted. \emph{Right:} the final posterior on the subhalo count correspondingly moves from a strong undershoot at $N=4$ to bracketing the true value (dashed line) at $N=20$.}
\label{fig:subhalo_N_scaling}
\end{figure}

If this explanation is right, the CDM undershoot should disappear when the data is made more informative. We test this directly by rerunning a single CDM run with $N=20$ lensing systems per evaluation, all other settings fixed (\cref{fig:subhalo_N_scaling}). The smooth-null share of the Round-1 posterior collapses from $96\%$ at $N=4$ to $4\%$ at $N=20$; the truth-covering abundant-NFW candidate rises from $0.2\%$ to $44\%$; and the final count posterior moves from $\hat n_{\rm mean} = 4 \pm 2$ to $36 \pm 10$, bracketing the true $35$. Note that the Round-1 null share measures information content rather than the final verdict: even at $N=4$, SMC refinement eventually displaces the null with near-degenerate \emph{ultra-sparse} NFW variants, which is what yields $\hat n_{\rm mean}=4\pm2$ rather than a smooth-lens call.

The family misidentifications are analogous. At $n_{\rm mean}=3$, a sparse cored population and a cored/NFW mixture produce nearly identical four-image statistics, so the narrower-prior mixed candidate wins by Occam on all three WDM seeds; the single SIDM miss likewise selects a mixed-profile candidate via the same cored/mixed near-degeneracy.

For an end-user the take-away is operational, i.e. the posterior is honest relative to the proposed candidate pool and the information content of the data, with the failure modes themselves being a diagnostic. An application of this framework to a CDM-like system at $N=4$ should read the narrow-prior preference and the prior-hugging count estimate as a signal that more lenses or higher-resolution imaging are needed, not as a confident claim of a sparse or smooth system.

\section{Discussion and Conclusion}
\label{sec:discussion}

We cast scientific model discovery as Bayesian inference in program/simulator space: an LLM supplies an implicit prior over simulators, and neural ratio estimation filters its samples into a joint posterior over models and parameters. The two components have complementary failure modes: LLMs generate plausible but sometimes physically incorrect models, while simulation-based inference provides rigorous comparison but needs candidates to compare. Because the ratio estimators target marginal likelihoods, complexity is penalized automatically: simpler models win ties by spreading their prior mass less thinly. Across our benchmarks the method recovers plausible model families from open-ended prompts.

A secondary property of this framework is that the posterior spreads across models when the data cannot distinguish candidates, rather than collapsing to a confident wrong answer. The lensing benchmark makes this concrete: the apparent CDM undershoot at $N{=}4$ is a genuine information-content limit, and in the single-seed rerun of \cref{fig:subhalo_N_scaling}, increasing the number of lensing systems per evaluation to $N{=}20$ eliminates the bias and recovers the truth. A corollary worth emphasizing is that the same Occam penalty that automatically rewards simpler models can, at low signal-to-noise, become the dominant factor in selection. This is Bayes-correct behavior in an information-limited setting, but practitioners should read it as a diagnostic of data uninformativeness rather than as evidence against broader or more complex models.

Several limitations remain. LLM-generated programs can pass validation while implementing incorrect physics, so domain expertise remains essential for sanity-checking outputs. The LLM also specifies each candidate's prior alongside its program, so downstream model comparison conflates ``which program fits the data'' with ``which prior matches the data'', a property we exploit (narrower well-placed priors are rewarded via marginal-likelihood Occam) but which means a user testing a specific scientific hypothesis should verify that the LLM-proposed priors actually bracket it. The NPE stage can also be loose for heavily mutated candidates with many weakly identified parameters (an intermediate $N{=}10$ subhalo rerun's parameter stage did not converge on its long-tailed within-NFW parameter space; the posterior-predictive check in \cref{fig:damped} therefore uses the exact-likelihood posterior available under the known noise model); model \emph{selection} is unaffected, since it rests on the evidence networks rather than the NPE. All benchmarks here are ``in-distribution'' for the LLM; genuinely out-of-distribution problems, where training data provides no relevant priors, remain to be explored. Finally, pairwise ratio estimation scales as $\binom{K}{2}$, so amortized ratio estimation that generalizes across models is a promising direction for larger candidate pools.

Treating model discovery as a probabilistic primitive alongside parameter inference expands the scope of simulation-based science from fitting a fixed model to finding which models the data supports. 
Because the inference stage is agnostic to the source of candidates, it evaluates whatever models are proposed, and as frontier LLMs continue to advance in code generation, hypothesis generation, and scientific reasoning, the proposal prior sharpens and the framework should inherit those gains directly, without methodological changes. Scientific model discovery in this framework can therefore ride the same capability curve as general-purpose LLM progress.

\section*{Acknowledgments}

We thank Carolina Cuesta-L\'azaro, Lukas Heinrich, and Michael Kagan for helpful conversations.
This research was supported by the Munich Institute for Astro-, Particle and BioPhysics (MIAPbP), which is funded by the Deutsche Forschungsgemeinschaft (DFG, German Research Foundation) under Germany's Excellence Strategy -- EXC-2094 -- 390783311.
The authors are pleased to acknowledge that the computational work reported on in this paper was performed on the Shared Computing Cluster which is administered by Boston University's Research Computing Services (\url{www.bu.edu/tech/support/research/}).
This work used PSC Bridges-2 GPU at the Pittsburgh Supercomputing Center through allocation PHY230064 from the Advanced Cyberinfrastructure Coordination Ecosystem: Services \& Support (ACCESS) program, which is supported by U.S. National Science Foundation grants \#2138259, \#2138286, \#2138307, \#2137603, and \#2138296.
\paragraph*{Software}

This work made use of \texttt{NumPy} \citep{harris2020array}, \texttt{SciPy} \citep{virtanen2020scipy}, \texttt{PyTorch} \citep{paszke2019pytorch}, \texttt{Matplotlib} \citep{Hunter:2007}, and the \texttt{sbi} package \citep{tejero2020sbi}.

\paragraph*{Use of Large Language Models}

Beyond its central role as the proposal model studied in the experiments, Claude Opus (versions~4.5--4.7) was used extensively throughout the preparation of this work: for ideation and discussion of methodology, for setting up and running experiments and analyses, and for drafting and revising the manuscript. Claude Fable~5 assisted with final revisions. The author takes full responsibility for the contents of the paper.

\bibliographystyle{plainnat}
\bibliography{references}

\newpage
\appendix

\section{Prompt Templates}
\label{app:prompts}

The generation prompt template is shown below; placeholders in braces are filled at run time with the benchmark's problem description, optional domain context, and constraints.
The prompt is structured to elicit diverse candidates while ensuring consistent output format.

\begin{lstlisting}[language={}]
You are a scientific modeling expert. Generate {n} candidate mathematical models for the following problem.

## Problem Description
{problem_description}

## Domain Context
{domain_context}

## Constraints
{constraints_text}

## Output Format
For each candidate model, provide:

1. A descriptive name
2. A brief description of the physical/mathematical motivation
3. Python code implementing a `simulate` function
4. Parameter specifications with suggested prior ranges

Use this exact JSON format (output ONLY valid JSON, no markdown):

```json
[
  {
    "name": "model_name",
    "description": "Brief description of the model and its motivation",
    "code": "def simulate(params, **kwargs):\n    # Implementation\n    return observations",
    "parameters": {
      "param1": {"type": "float", "prior_type": "uniform", "prior_min": 0.0, "prior_max": 1.0, "description": "Description"},
      "param2": {"type": "float", "prior_type": "loguniform", "prior_min": 1e-3, "prior_max": 1e3, "description": "Description"}
    }
  }
]
```

CRITICAL Requirements for the `simulate` function:
- Takes a dict `params` containing parameter values
- May take additional kwargs (e.g., `t` for time points, `n_obs` for number of observations)
- Returns numpy array of simulated observations
- numpy is ALREADY imported as `np` - DO NOT include any import statements
- DO NOT use `import numpy` or `from scipy` or any imports - just use `np` directly
- Should be numerically stable
- Use only numpy functions (np.exp, np.sin, np.cos, np.sqrt, etc.)

IMPORTANT: Simulators must include OBSERVATIONAL NOISE!
- Get noise level from kwargs: `noise = kwargs.get('noise_std', 0.1)`
- Add Gaussian noise to the output: `return signal + noise * np.random.randn(len(signal))`
- This is critical for proper Bayesian model comparison - the simulator must match the data-generating process

Generate {n} diverse candidates that explore different modeling assumptions.

IMPORTANT: Include a range of model complexities:
- At least one MINIMAL model with only 2-3 parameters (the simplest plausible explanation)
- Some intermediate complexity models (3-4 parameters)
- At most one complex model (5+ parameters) if truly justified

Simpler models are preferred if they can explain the data. Start simple.
\end{lstlisting}

When the optional fields are empty, the literal fallbacks ``No additional context provided.'' and ``No specific constraints.'' are substituted for \texttt{\{domain\_context\}} and \texttt{\{constraints\_text\}}.
Note that the template's closing block explicitly requests a range of candidate complexities, including at least one minimal model, and states a preference for simpler models. This shapes the \emph{proposal} distribution $p_{\text{LLM}}$, i.e.\ which programs enter the pool; the complexity penalty applied during model \emph{comparison} remains the marginal-likelihood Occam mechanism of Section~\ref{sec:ratio} alone.

\section{Additional Experimental Details}
\label{app:experiments}

\paragraph{Neural Network Architectures}
For ratio estimation, we use 4-layer MLPs with 256 hidden units and GELU activations.
Each evidence network is trained with the exponential loss $\mathcal{L}(f, m) = \exp((0.5 - m) \cdot J_2(f))$ (with $\ell$-POP transform $J_2(f) = f(1+|f|)$), Adam optimizer (learning rate $10^{-3}$), 80/20 train/validation split, and early stopping after four epochs without validation improvement. For the ODE benchmarks the evidence networks operate on 32-dimensional learned summary statistics produced by a summary network trained per round, rather than on the raw time series.
We train ensembles of 5 networks per model pair to reduce variance.
For the subhalo benchmark, the evidence networks operate on 128-dimensional set embeddings produced by a ResNet-18 image encoder followed by a DeepSets permutation-invariant pool (Section~\ref{sec:set-level}). The set embedder is trained to classify the current $K$-model pool and is retrained once per SMC round so its feature space tracks mutated models.

For neural posterior estimation, we use masked autoregressive flows (MAFs) from the \texttt{sbi} package, trained on (parameter, observation) pairs drawn from each surviving model's prior; flows are trained for the top three surviving models in the subhalo benchmark and for the top-ranked model in the ODE benchmarks.
Training uses the Adam optimizer at the \texttt{sbi} default learning rate ($5 \times 10^{-4}$) for up to 200 epochs; the SBC and parameter-recovery studies in Appendix~\ref{app:sbc} use a learning rate of $10^{-4}$.

\paragraph{Simulation Parameters}
\emph{Damped oscillator:} 50 uniformly spaced time points over $t \in [0, 10]$; observation noise $\sigma = 0.1$; priors $A \sim \text{Uniform}(0.5, 5)$, $\gamma \sim \text{Uniform}(0.1, 1.5)$, $\omega \sim \text{Uniform}(1, 6)$.

\emph{SIR:} Gillespie simulation with $\beta = 0.4$, $\gamma = 0.1$, population $N = 1000$, initial infected $I_0 = 10$; 10 observations over $t \in [0, 100]$ with noise $\sigma_\text{obs} = 5$; priors are LLM-specified per candidate, centered on physically plausible rates.

\emph{Subhalo:} 64$\times$64 pixel images with field of view $\pm$3 arcsec; main singular-isothermal-ellipsoid (SIE) lens with $\theta_E\in[1.3,1.8]$ arcsec, $q\in[0.7, 0.95]$, source offset $\sim\mathcal{N}(0, 0.05)$ arcsec (all varied per image); subhalo masses drawn from a power-law mass function in the range $[10^7, 10^{10}] M_\odot$; density profiles NFW (cuspy), cored, or a 50/50 mixture; 2-channel input (image + residual); ResNet-18 image encoder + DeepSets set pool producing a single 128-dimensional embedding per 4-image set; ensembles of 5 evidence networks per model pair with $\ell$-POP exponential loss; 320 training datasets per candidate model, each dataset containing $N=4$ images sharing one population-parameter vector; 4-image hierarchical set-level evaluation at test time; set embedder retrained at each SMC round; force-inclusion of a smooth-null model in the Round-1 pool; a minimum of 3 SMC rounds before the weight-based termination can trigger; final-round model weights computed on a held-out scoring split, disjoint from the images used during search; 3 random seeds per ground truth across 6 ground truths (CDM, CDM-low, WDM, WDM-high, SIDM, Mixed); up to $R=15$ refinement rounds; open-ended prompts; LLM: Claude Opus 4.6 (\texttt{claude-opus-4-6}).

\paragraph{Computational Resources}
All experiments were run on a heterogeneous pool of NVIDIA GPUs (predominantly L40S, plus A100 and RTX~PRO~6000; one GPU per run) via a shared scheduler.
Wall-clock times: ODE benchmarks $\sim$2--30 minutes per run; subhalo runs $\sim$45--70 minutes per ground-truth/seed pair (median ${\sim}$53), dominated by per-round retraining of the 200-epoch set embedder (ResNet-18 + DeepSets) across up to 15 SMC rounds.
The full subhalo experiment is 6 ground truths $\times$ 3 seeds $=$ 18 runs, totalling ${\sim}$16 GPU-hours of compute, run in parallel across several GPUs via the shared scheduler.
Per-round work breaks down as: set-embedder training (ResNet-18 + DeepSets, 200 epochs), pairwise evidence-network training ($\binom{K}{2}$ pairs $\times$ 5 ensemble members), and a single MAF training per surviving top-ranked model for parameter inference.

\section{Least-Squares Aggregation: Derivations}
\label{app:ls_aggregation}

This appendix gives the two derivations referenced in Section~\ref{sec:ratio}: the implicit inverse temperature of naive averaging, and the closed-form solution of the least-squares aggregation problem.

\begin{figure}[t]
\centering
\includegraphics[width=\textwidth]{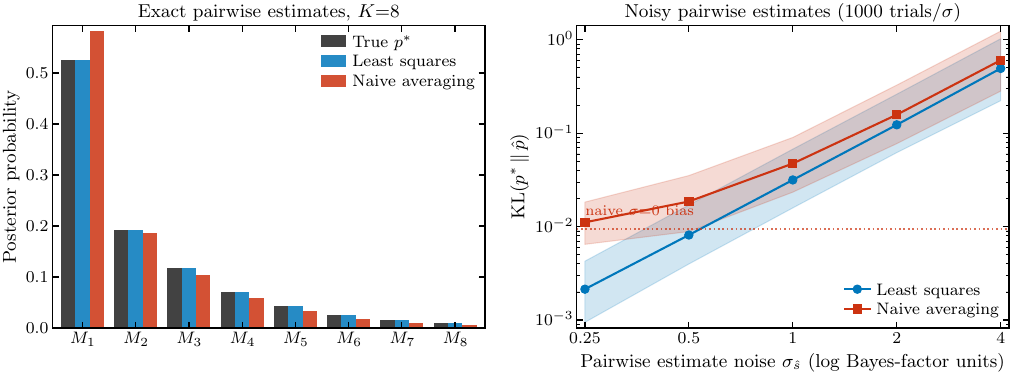}
\caption{Aggregating pairwise log Bayes factors: least squares recovers the true posterior where naive averaging sharpens it. Synthetic data: $K{=}8$ models with hand-chosen true log-evidences $\boldsymbol{\ell}^{\,*}$. \emph{(a)} Exact pairwise estimates $\hat{s}_{ij} = \ell^*_i - \ell^*_j$: \textcolor[HTML]{0077BB}{Least squares} recovers the true posterior $p^*_k \propto \exp(\ell^*_k)$ to machine precision; \textcolor[HTML]{CC3311}{naive averaging} produces softmax of $\tfrac{K}{K-1}\,\boldsymbol{\ell}^{\,*}$ exactly, a sharpened posterior with the leading-model weight inflated by ${\sim}11\%$ (relative). \emph{(b)} Independent Gaussian noise of std $\sigma_{\hat s}$ added to each pairwise estimate (1000 trials per $\sigma$); curve is the median, band is 16--84 percentile. Least squares is strictly better than naive averaging at every noise level tested, and at small noise it is far better: naive is bounded below by the $\sigma{=}0$ temperature-distortion floor (dotted line) while least squares scales smoothly from $0$.}
\label{fig:ls_synthetic}
\end{figure}

\paragraph{Naive averaging has implicit inverse temperature $K/(K{-}1)$.}
Assume exact pairwise estimates $\hat{s}_{ij} = \ell_i - \ell_j$ for true log-evidences $\ell_1, \ldots, \ell_K$. The naive aggregator computes, for each model $k$,
\begin{equation}
    \bar{s}_k \;\equiv\; \frac{1}{K-1} \sum_{j \neq k} \hat{s}_{kj}
              \;=\; \frac{1}{K-1} \sum_{j \neq k} (\ell_k - \ell_j)
              \;=\; \frac{K}{K-1}\,\ell_k \;-\; \frac{1}{K-1}\sum_{j=1}^{K} \ell_j.
\end{equation}
The second term is the same constant for every $k$, so it drops out under softmax. The first term, however, scales the true log-evidences by $K/(K{-}1)$. The softmax-induced posterior is therefore
\begin{equation}
    p_{\text{naive}}(M_k \given \obs) \;\propto\; \exp\!\left(\tfrac{K}{K-1}\,\ell_k\right),
\end{equation}
which is the true Bayesian posterior raised to the inverse temperature $\beta = K/(K{-}1) > 1$. At our $K{=}8$ this is $\beta = 8/7 \approx 1.14$, so the largest posterior weight is artificially inflated and the smallest weights artificially suppressed (see \cref{fig:ls_synthetic} for the numerical consequence). The bias does not vanish in any limit of the data: it is a property of the aggregation rule, present whenever the naive average is composed with a softmax to define the model posterior.

\paragraph{Least-squares aggregation as a Laplacian solve.}
Stack the $n = \binom{K}{2}$ observations into a vector $\mathbf{b} \in \mathbb{R}^{n}$ with entries $b_{(i,j)} = \hat{s}_{ij}$ for $i < j$. Let $\mathbf{A} \in \{-1, 0, +1\}^{n \times K}$ be the corresponding pairwise design matrix, with row $(i,j)$ containing $+1$ at column $i$, $-1$ at column $j$, and zeros elsewhere. The objective in \cref{eq:ls_aggregation} is $\|\mathbf{A}\boldsymbol{\ell} - \mathbf{b}\|_2^2$, whose normal equations are
\begin{equation}
    \mathbf{A}^\top \mathbf{A}\,\boldsymbol{\ell} \;=\; \mathbf{A}^\top \mathbf{b}.
\end{equation}
Direct computation gives $\mathbf{A}^\top \mathbf{A} = K\,\mathbf{I} - \mathbf{1}\mathbf{1}^\top$, which is the unnormalized graph Laplacian of the complete graph on $K$ vertices. This matrix has eigenvalues $K$ with multiplicity $K-1$ and $0$ with multiplicity $1$ (eigenvector $\mathbf{1}$, reflecting that $\boldsymbol{\ell}$ is only identified up to an additive constant). The Moore--Penrose pseudoinverse is
\begin{equation}
    (\mathbf{A}^\top \mathbf{A})^{+} \;=\; \tfrac{1}{K}\left(\mathbf{I} - \tfrac{1}{K}\mathbf{1}\mathbf{1}^\top\right),
\end{equation}
which directly enforces the gauge $\sum_k \ell_k = 0$. The solution is $\boldsymbol{\ell}^{\,*} = (\mathbf{A}^\top \mathbf{A})^{+}\,\mathbf{A}^\top\mathbf{b}$, which we compute equivalently via a singular value decomposition (SVD) of $\mathbf{A}$ using \texttt{scipy.linalg.lstsq} (our implementation uses the alternative gauge $\ell_0 = 0$ by removing model~$0$ from the design matrix; both gauges give the same posterior and differ only by an additive constant in $\boldsymbol{\ell}$). On the complete comparison graph the solution has a simple closed form, $\ell^{*}_k = \frac{1}{K}\sum_{j} \hat{s}_{kj}$ with $\hat{s}_{kk}=0$: the row mean over all $K$ entries, \emph{including} the trivially consistent self-comparison. This makes the relationship to naive averaging transparent: least squares produces the same ranking of models but removes the self-exclusion bias responsible for the $K/(K{-}1)$ temperature, and additionally exposes the cyclic residual below.

Two properties follow from this construction. First, when $\mathbf{b}$ is consistent (i.e.\ $b_{(i,j)} = \ell^*_i - \ell^*_j$ for some $\boldsymbol{\ell}^{\,*}$), the least-squares solution recovers $\boldsymbol{\ell}^{\,*}$ exactly up to the gauge; there is no implicit temperature distortion. Second, when $\mathbf{b}$ is inconsistent (cyclic), the solution is the orthogonal projection of $\mathbf{b}$ onto the image of $\mathbf{A}$ in $L^2$: geometrically, the closest consistent comparison matrix to the observed one. The residual $\|\mathbf{A}\boldsymbol{\ell}^{\,*} - \mathbf{b}\|_2$ is a directly computable measure of how non-transitive the ensemble of pairwise classifiers is on the test observation; it is available as a diagnostic but is not propagated into the posterior. This perspective is the Hodge decomposition of pairwise comparison data: the image of $\mathbf{A}$ is the ``gradient'' subspace (consistent with a global potential $\boldsymbol{\ell}$), its orthogonal complement is the ``curl'' subspace of purely cyclic residuals (on the complete comparison graph the harmonic component of the Hodge decomposition vanishes), and least squares cleanly separates the two \citep{jiang2011statistical}.

\paragraph{Synthetic illustration.}
\Cref{fig:ls_synthetic} verifies both derivations directly. We fix $K{=}8$ models with hand-chosen true log-evidences $\boldsymbol{\ell}^{\,*}$ and compare three quantities: the true posterior $p^*_k \propto \exp(\ell^*_k)$, the least-squares posterior, and the naive-averaged posterior. With \emph{exact} pairwise estimates $\hat{s}_{ij} = \ell^*_i - \ell^*_j$ (panel a), the least-squares posterior reproduces $p^*$ to machine precision, while the naive posterior matches the predicted softmax of $\beta\,\boldsymbol{\ell}^{\,*}$ with $\beta = K/(K{-}1) \approx 1.14$: the leading model's weight is inflated from $0.52$ to $0.58$, with compensating suppression of the tail. With Gaussian noise added independently to each pairwise estimate (panel b), the least-squares posterior's Kullback--Leibler (KL) divergence from $p^*$ rises smoothly from zero with $\sigma_{\hat s}$, while the naive posterior is bounded below by its $\sigma{=}0$ temperature-distortion floor (dotted line) and only catches up to least squares at large noise where both methods degrade. The simulation involves no LLM, GPU, or benchmark data; it isolates the aggregation step.

\section{Generated Model Examples}
\label{app:models}

Below are the winning LLM-generated simulator programs from three damped-oscillator runs that converged in Round~1.

\paragraph{\texttt{damped\_cosine}}
\begin{lstlisting}[language=Python]
def simulate(params, **kwargs):
    t = kwargs.get('t', np.linspace(0, 10, 50))
    noise = kwargs.get('noise_std', 0.1)
    seed = kwargs.get('seed', None)
    if seed is not None:
        np.random.seed(seed)
    A = params['A']
    gamma = params['gamma']
    omega = params['omega']
    signal = A * np.exp(-gamma * t) * np.cos(omega * t)
    return signal + noise * np.random.randn(len(t))
\end{lstlisting}

\paragraph{\texttt{gaussian\_envelope\_cosine}}
\begin{lstlisting}[language=Python]
def simulate(params, **kwargs):
    t = kwargs.get('t', np.linspace(0, 10, 50))
    noise = kwargs.get('noise_std', 0.1)
    seed = kwargs.get('seed', None)
    if seed is not None:
        np.random.seed(seed)
    A = params['A']
    sigma = params['sigma']
    omega = params['omega']
    signal = A * np.exp(-(t / sigma) ** 2) * np.cos(omega * t)
    return signal + noise * np.random.randn(len(t))
\end{lstlisting}

\paragraph{\texttt{power\_law\_decay\_cosine}}
\begin{lstlisting}[language=Python]
def simulate(params, **kwargs):
    t = kwargs.get('t', np.linspace(0, 10, 50))
    noise = kwargs.get('noise_std', 0.1)
    seed = kwargs.get('seed', None)
    if seed is not None:
        np.random.seed(seed)
    A = params['A']
    alpha = params['alpha']
    omega = params['omega']
    envelope = A / (1.0 + t) ** alpha
    signal = envelope * np.cos(omega * t)
    return signal + noise * np.random.randn(len(t))
\end{lstlisting}

The LLM produces syntactically correct code with a common keyword-argument interface (time grid, noise level, seed) supplied by the harness.
The three programs implement distinct envelope hypotheses (exponential, Gaussian, and power-law decay), enabling meaningful model comparison.

\section{Simulation-Based Calibration}
\label{app:sbc}

We assess posterior calibration using simulation-based calibration (SBC) \citep{talts2018validating}.
The SBC targets the NPE machinery on the damped-cosine simulator $x(t) = A e^{-\gamma t} \cos(\omega t)$ with uniform priors $A \in [0.5, 5]$, $\gamma \in [0.05, 1.5]$, $\omega \in [1, 6]$, observation noise $\sigma = 0.15$, and the same MAF architecture and training configuration as the main experiments.
For each of 200 test cases, we: (1) sample true parameters $\params^*$ from the prior, (2) generate synthetic observation $\obs^* \sim p(\cdot|\params^*, M)$, (3) draw 1000 posterior samples $\{\params^{(i)}\}$ from $q_\psi(\params|\obs^*)$, and (4) compute the rank of $\params^*$ among the posterior samples for each parameter dimension.

For well-calibrated posteriors, ranks should be uniformly distributed over $\{1, \ldots, 1001\}$.
Systematic biases appear as non-uniform histograms: under-coverage (overconfident posteriors) produces U-shaped histograms, while over-coverage (overly conservative posteriors) produces peaked histograms.

\begin{figure}[h]
\centering
\includegraphics[width=0.65\columnwidth]{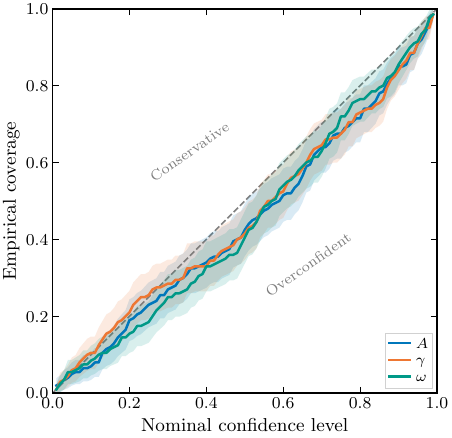}
\caption{Coverage calibration for the damped oscillator benchmark (200 SBC test cases): the NPE posteriors are approximately calibrated, with mild under-coverage. $x$-axis: nominal confidence level $\alpha$; $y$-axis: empirical coverage, the fraction of test cases where the true parameter falls within the $\alpha$-credible interval of the NPE posterior. Curves for the three parameters ($A$, $\gamma$, $\omega$) sit slightly below the diagonal, indicating mild under-coverage (the NPE posteriors are slightly tighter than nominal). Shaded bands are 95\% binomial confidence intervals on the empirical coverage estimates.}
\label{fig:coverage}
\end{figure}

\paragraph{Coverage Statistics}
\Cref{fig:coverage} shows the coverage calibration for the damped oscillator benchmark computed from 200 SBC test cases, plotting empirical coverage against nominal confidence level for each of the three parameters ($A$, $\gamma$, $\omega$).
A well-calibrated posterior tracks the diagonal; points above indicate conservative posteriors (wider than necessary), while points below indicate overconfidence.

Empirical coverage sits slightly below nominal across the range tested, consistent across all three parameters. This indicates mildly overconfident posteriors (the NPE assigns somewhat tighter credible intervals than the true rate implies), but the effect is modest and the ranking of inferred parameters is unaffected. This SBC certifies the NPE machinery in the well-specified setting, where training and test data come from the same simulator; it does not certify posteriors under the pipeline's LLM-generated, possibly misspecified candidates.

\paragraph{Point-Estimate Recovery (Representative Run)}
For a concrete illustration of the posterior's width and location, \cref{tab:params_damped} lists the marginal mean, standard deviation, and 90\% credible interval of each parameter for the stretched-exponential oscillator candidate, a recurring LLM proposal across runs that nests the ground truth, using an NPE trained at the benchmark settings ($\sigma = 0.1$, 5{,}000 simulations).
All 90\% intervals cover the ground truth, consistent with the SBC result; the envelope parameters $\tau$ and $\beta$ are weakly identified from a single observation and correspondingly broad. Coverage of all four parameters is stable across repeated NPE retrainings and across observation seeds, though the interval widths vary between retrainings (the weakly identified $\tau$ most of all); the table reports the committed, seeded run.

\begin{table}[h]
\centering
\caption{Parameter recovery for the stretched-exponential oscillator candidate $A \exp(-(t/\tau)^\beta) \cos(\omega t)$, a recurring LLM-proposed family that nests the ground truth $x(t) = A \exp(-\gamma t) \cos(\omega t)$ ($A{=}2.0$, $\gamma{=}0.3$, $\omega{=}3.0$) at $\beta{=}1$, $\tau{=}1/\gamma{=}3.33$. NPE trained on 5{,}000 simulations at the benchmark noise $\sigma = 0.1$; posterior for one observation generated at the ground truth. All 90\% credible intervals cover the truth; the envelope parameters $\tau$ and $\beta$ trade off against each other and are weakly identified from a single observation, reflected in their broad intervals.}
\label{tab:params_damped}
\sisetup{table-format=1.2, table-number-alignment=center}
\begin{tabular}{l S S S c}
\toprule
{Parameter} & {True} & {Mean} & {Std} & {90\% CI} \\
\midrule
Amplitude $A$ & 2.00 & 2.07 & 0.10 & {[1.92, 2.25]} \\
Decay $\tau$ & 3.33 & 3.87 & 1.52 & {[1.40, 6.49]} \\
Stretch $\beta$ & 1.00 & 1.66 & 0.62 & {[0.60, 2.70]} \\
Frequency $\omega$ & 3.00 & 3.02 & 0.03 & {[2.97, 3.07]} \\
\bottomrule
\end{tabular}
\end{table}

\end{document}